\documentclass[11pt]{article}

\usepackage[final]{automl}

\usepackage{microtype} 
\usepackage{booktabs}  
\usepackage{url}  
\usepackage{csquotes}
\usepackage{natbib}

\usepackage[utf8]{inputenc}
\usepackage{url}

\usepackage{amsmath}
\usepackage{bbm}
\usepackage{bm}
\usepackage{cancel}
\usepackage{mathtools}

\usepackage{ascmac}  
\usepackage{algorithm}
\usepackage{algpseudocode}

\usepackage{fancyhdr}
\usepackage{tcolorbox}
\usepackage{tikz}

\usepackage{booktabs}
\usepackage{comment}
\usepackage{lastpage}
\usepackage{listings}
\usepackage{multibib}
\usepackage{multirow}
\usepackage[super]{nth}

\tcbuselibrary{breakable, skins, theorems}

\allowdisplaybreaks

\algtext*{EndFor}
\algtext*{EndWhile}
\algtext*{EndIf}
\algtext*{EndProcedure}
\algtext*{EndFunction}
\algnewcommand{\LineComment}[1]{\State \(\triangleright\) #1}

\newcites{appx}{References}

\newif\ifunderreview
\newif\ifsubmission
\newif\ifappendix

\underreviewfalse

\submissiontrue

\appendixtrue

\ifsubmission
\newcommand{\todo}[1]{}
\newcommand{\replace}[2]{}
\newcommand{\sw}[1]{}

\else
\newcommand{\todo}[1]{\textbf{\textcolor{red}{[TODO: #1]}}}

\newcommand{\replace}[2]{\textbf{\textcolor{red}{[del: \cancel{#1}]}}\textbf{\textcolor{blue}{[new: #2]}}}

\newcommand{\sw}[1]{\textbf{\textcolor{cyan}{[SW: #1]}}}

\usepackage[inline]{showlabels}

\fi

\makeatletter
\newcommand{\customlabel}[2]{%
   \protected@write \@auxout {}{\string \newlabel {#1}{{#2}{\thepage}{#2}{#1}{}} }%
   \hypertarget{#1}{}
}
\makeatother

\ifunderreview
\hypersetup{%
  pdfauthor={Paper under double-blind review}, 
  pdftitle={Python Wrapper for Simulating Asynchronous Optimization on Cheap Benchmarks without Any Wait},
  pdfsubject={async HPO simulator},
  pdfkeywords={Python Tool, Asynchronous optimization, Multi-fidelity optimization}
}
\else
\hypersetup{%
  pdfauthor={Shuhei Watanabe}, 
  pdftitle={Python Wrapper for Simulating Asynchronous Optimization on Cheap Benchmarks without Any Wait},
  pdfsubject={async HPO simulator},
  pdfkeywords={Python Tool, Asynchronous optimization, Multi-fidelity optimization}
}

\fi

\newcommand{\argmin}{\mathop{\mathrm{argmin}}}

\renewcommand{\eqref}[1]{Eq.~(\ref{#1})}

\newcommand{\D}{\mathcal{D}}
\newcommand{\X}{\mathcal{X}}
\newcommand{\I}{\mathcal{I}}
\newcommand{\av}{\boldsymbol{a}}
\newcommand{\xv}{\boldsymbol{x}}
\newcommand{\zv}{\boldsymbol{z}}

\DeclareFixedFont{\ttb}{T1}{txtt}{bx}{n}{12} 
\DeclareFixedFont{\ttm}{T1}{txtt}{m}{n}{12}  

\definecolor{deepblue}{rgb}{0,0,0.5}
\definecolor{deepred}{rgb}{0.6,0,0}
\definecolor{deepgreen}{rgb}{0,0.5,0}

\begin{document}
\title{Python Wrapper for Simulating Multi-Fidelity Optimization on HPO Benchmarks without Any Wait}

\ifunderreview
\author{
  Paper under double-blind review \\
}
\else
\author{
  Shuhei Watanabe \\
  Department of Computer Science, University of Freiburg, Germany\\
  \texttt{watanabs@cs.uni-freiburg.de}
}
\fi

\maketitle

\begin{abstract}
  Hyperparameter (HP) optimization of deep learning (DL) is essential for high performance.
  As DL often requires several hours to days for its training, HP optimization (HPO) of DL is often prohibitively expensive.
  This boosted the emergence of tabular or surrogate benchmarks, which enable querying the (predictive) performance of DL with a specific HP configuration in a fraction.
  However, since the actual runtime of a DL training is significantly different from its query response time, simulators of an asynchronous HPO, e.g. multi-fidelity optimization, must wait for the actual runtime at each iteration in a na\"ive implementation; otherwise, the evaluation order during simulation does not match with the real experiment.
  To ease this issue, we developed a Python wrapper and describe its usage.
  This wrapper forces each worker to wait so that we yield exactly the same evaluation order as in the real experiment with only $10^{-2}$ seconds of waiting instead of waiting several hours.
  Our implementation is available at
  \ifunderreview
  \url{https://anonymous.4open.science/r/mfhpo-simulator-3C81/}.
  \else
  \url{https://github.com/nabenabe0928/mfhpo-simulator/}.
  \fi
\end{abstract}

\section{Introduction}
Hyperparameter (HP) optimization of deep learning (DL) is crucial for strong performance~(\cite{chen2018bayesian,henderson-aaai18a}) and it surged the research on HP optimization (HPO) of DL.
However, due to the heavy computational nature of DL, HPO is often prohibitively expensive and both energy and time costs are not negligible.
This is the driving force behind the emergence of tabular and surrogate benchmarks, which enable yielding the (predictive) performance of a specific HP configuration in less than a second~(\cite{eggensperger2015efficient,eggensperger2021hpobench,arango2021hpo,bansal2022jahs}).

Although these benchmarks effectively reduce the energy usage and the runtime of experiments in many cases, experiments considering runtimes between parallel workers may not be easily benefited.
For example, multi-fidelity optimization (MFO)~(\cite{kandasamy2017multi}) has been actively studied recently due to its computational efficiency~(\cite{jamieson2016non,li2017hyperband,falkner2018bohb,awad2021dehb}), but because of its asynchronous nature, the call order of each worker must be appropriately sorted out to not break the states that the actual experiments would go through.
While this problem is na\"ively addressed by making each worker wait for the runtime the actual DL training would take, each worker must wait for a substantial amount of time in this case, and hence it ends up wasting energy and time.

To address this problem, we developed a Python wrapper that automatically sorts out both the order of evaluations and the allocation of jobs to each worker in a fraction by forcing each worker to wait internally.
In this paper, we describe how our package solves the problem and how to use this wrapper.
Our package can be easily used for existing parallel optimizers such as BOHB~(\cite{falkner2018bohb}), DEHB~(\cite{awad2021dehb}), and SMAC3~(\cite{lindauer2022smac3}) and we provide the example codes of how to pass our wrapper to these existing optimizers at
\ifunderreview
\url{https://anonymous.4open.science/r/mfhpo-simulator-3C81/}.
\else
\url{https://github.com/nabenabe0928/mfhpo-simulator/}.
\fi

\section{Background}
In this section, we describe the potential problems during MFO on tabular or surrogate benchmarks and explain why our wrapper would be useful.
Throughout the paper, we assume minimization problems of an objective function $f(\xv): \X \rightarrow \mathbb{R}$ defined on the search space $\X \coloneqq \X_1 \times \X_2 \times \dots \times \X_D$ where $\X_d \subseteq \mathbb{R}$ is the domain of the $d$-th HP.
Furthermore, we define the (predictive) \emph{actual} runtime function $\tau(\xv): \X \rightarrow \mathbb{R}_{+}$ of the objective function given an HP configuration $\xv$.
Note that although $f(\xv)$ and $\tau(\xv)$ could involve randomness, we only describe the deterministic version for the notational simplicity, and we use the notation $[i] \coloneqq \{1,2,\dots,i\}$ where $i$ is a positive integer throughout this paper.

\subsection{Asynchronous Optimization on Surrogate or Tabular Benchmarks}
Assume we have the oracle, i.e. tabular or surrogate benchmark, of $f$ and $\tau$ that could be queried in a negligible amount of time, the HP configuration $\xv$ is sampled from a policy $\pi(\xv | \D^{(N)})$ where $\D^{(N)} \coloneqq \{(\xv_n, f(\xv_n))\}_{n=1}^{N}$ is a set of observations, and we have a set of parallel workers $\{W_p\}_{p=1}^P$ where each worker $W_p: \X \rightarrow \mathbb{R}^2$ is a wrapper of $f(\xv)$ and $\tau(\xv)$.
Let a mapping $i_n: \mathbb{Z}_{+}\rightarrow [P]$ be an index specifier of which worker processed the $n$-th observation and $\I_p^{(N)} \coloneqq \{n \in [N] \mid i_n = p\}$ be a set of indices of observations the $p$-th worker processed.
Then the (simulated) runtime of the $p$-th worker is computed as follows if we ignore the sampling time:
\begin{equation}
  \begin{aligned}
    T_p^{(N)} \coloneqq \sum_{n \in \I_p^{(N)}} \tau(\xv_n).
  \end{aligned}
\end{equation} 
In turn, the ($N + 1$)-th observation will be processed by the worker that will be free for the first time, and thus the index of the worker for the ($N + 1$)-th observation is specified by $\argmin_{p \in [P]} T_p^{(N)}$.

\begin{figure}[tb]
  \begin{center}
    \subfloat[\sloppy Wrapper Workflow\label{main:methods:subfig:api}]{
      \includegraphics[width=0.4\textwidth]{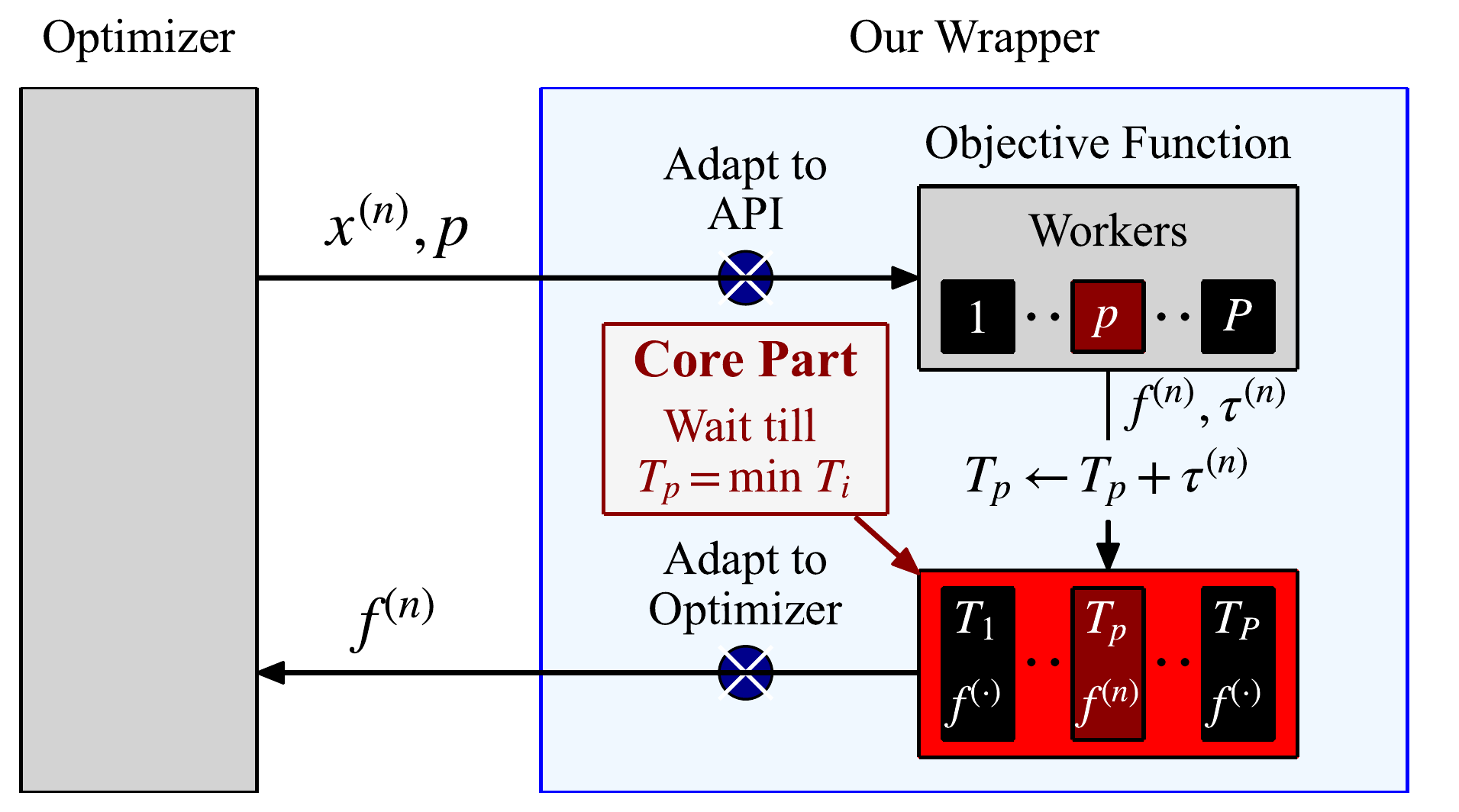}
    }
    \subfloat[\sloppy Compression of Simulated Runtime\label{main:methods:subfig:compress}]{
      \includegraphics[width=0.58\textwidth]{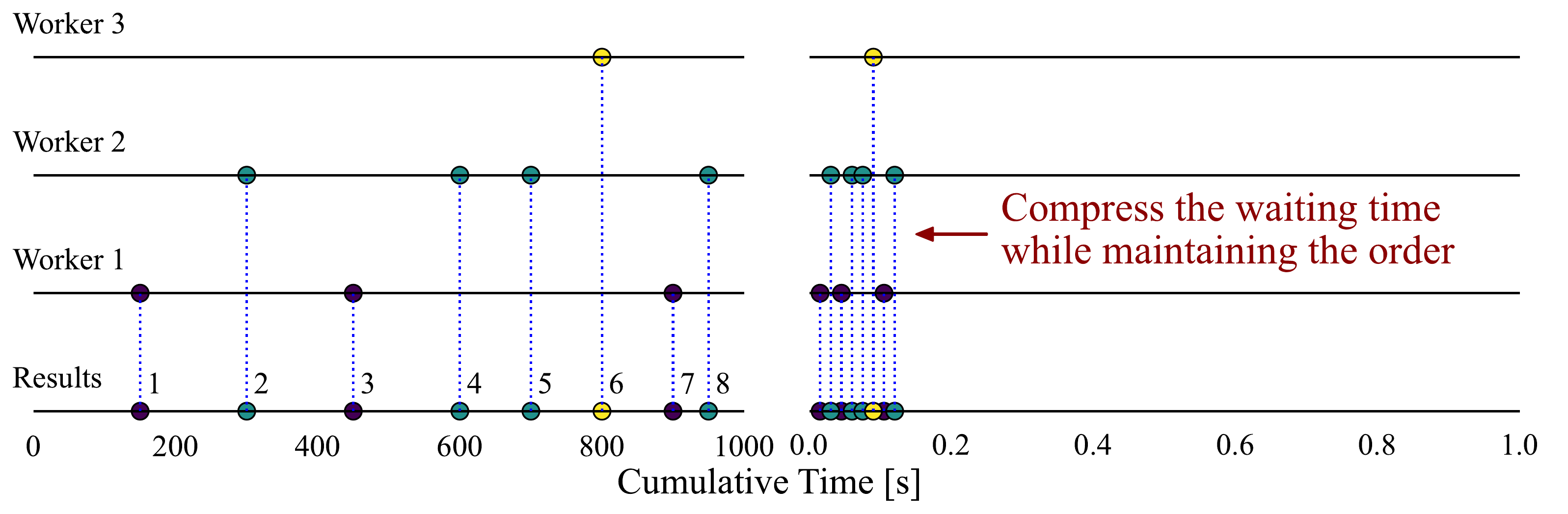}
    }
    \caption{
      The conceptual visualizations of our wrapper.
      (a) The workflow of our wrapper.
      The gray parts are provided by users and our package is responsible for the light blue part.
      The blue circles with the white cross must be modified by users to match the signature used in our wrapper via inheritance as described in Listings~\ref{main:usage:code:inheritance} and \ref{main:usage:code:non-mfo}.
      The $p$-th worker receives the $n$-th queried configuration $\xv^{(n)}$ and stores its result $f^{(n)}\coloneqq f(\xv^{(n)}), \tau^{(n)} \coloneqq \tau(\xv^{(n)})$ in the file system.
      Our wrapper sorts out the right timing to return the $n$-th queried result $f^{(n)}$ to the optimizer based on the simulated runtime $T_p$.
      (b) The compression of simulated runtime.
      Each circle on each line represents the timing when each result was delivered from each worker.
      \textbf{Left}: an example when we na\"ively wait for the (actual) runtime $\tau(\xv)$ of each query.
      Although we receive each result at the right timing, the simulation incurs a substantial amount of waiting time. 
      \textbf{Right}: an example when we use our wrapper.
      We can significantly reduce the experiment time without losing the exact order.
      \label{main:methods:fig:conceptual}
    }
  \end{center}
\end{figure}

The problems of this setting are that (1) the policy is conditioned on $\D^{(N)}$ and the order of the observations must be preserved and (2) each worker must wait for the other workers to match the order to be realistic.
While an obvious approach is to let each worker wait for the queried runtime $\tau(\xv)$ as seen in Figure~\ref{main:methods:subfig:compress}, it is a waste of energy and time.
Therefore, we developed a Python wrapper to match the order while each worker waits for only a \emph{negligible amount of time}.

\subsection{Related Work}
Although there have been many HPO benchmarks invented for MFO such as HPOBench~(\cite{eggensperger2021hpobench}), NASLib~(\cite{mehta2022bench}), and JAHS-Bench-201~(\cite{bansal2022jahs}), none of them provides a module to allow researchers to simulate runtime internally.
Other than HPO benchmarks, many HPO frameworks handling MFO have also been developed so far such as Optuna~(\cite{akiba2019optuna}), SMAC3~(\cite{lindauer2022smac3}), Dragonfly~(\cite{kandasamy2020tuning}), and RayTune~(\cite{liaw2018tune}).
However, no framework above considers the simulation of runtime.
Although HyperTune~(\cite{li2022hyper}) and SyneTune~(\cite{salinas2022syne}) are internally simulating the runtime, we cannot simulate optimizers of interest if the optimizers are not introduced in the packages.
It implies that researchers cannot immediately simulate their own new methods unless they incorporate their methods in these packages.
Furthermore, their simulation backend does not support multiprocessing and multithreading.
Therefore, an easy-to-use Python wrapper for the simulation is needed.

\section{Automatic Waiting Time Scheduling Wrapper}

As an objective function may take a random seed and fidelity parameters in practice, we denote a set of the arguments for the $n$-th query as $\av^{(n)}$.
Additionally, \emph{job} means to allocate the $n$-th queried HP configuration $\xv^{(n)}$ to a free worker and obtain its result $r^{(n)} \coloneqq (f(\xv^{(n)} | \av^{(n)}), \tau(\xv^{(n)} | \av^{(n)}))$.
Besides that, we denote the $n$-th chronologically ordered result as $r_n$.
Our wrapper is required to satisfy the following features:
\begin{itemize}
  \vspace{-1mm}
  \item The $i$-th result $r_i$ comes earlier than the $j$-th result $r_j$ for all $i < j$,
  \vspace{-1mm}
  \item The wrapper recognizes each worker and allocates a job to the exact worker even when using multiprocessing~(e.g. \texttt{joblib} and \texttt{dask}) and multithreading~(e.g. \texttt{concurrent.futures}),
  \vspace{-1mm}
  \item The evaluation of each job can be restarted in MFO.
  \vspace{-1mm}
\end{itemize}
Note that an example of the restart of evaluation means that when we already evaluate DL with $\xv$ for $20$ epochs and would like to evaluate DL with the same HP configuration $\xv$ for $100$ epochs, we start the training of DL from the $21$st epoch instead of from scratch using the intermediate state.
To achieve these features, we chose to share the required information via the file system and create the following JSON files that map:
\begin{itemize}
  \vspace{-1mm}
  \item from a thread or process ID of each worker to a worker index $p \in [P]$,
  \vspace{-1mm}
  \item from a worker index $p \in [P]$ to its (simulated) cumulative runtime $T_p^{(N)}$, and
  \vspace{-1mm}
  \item from the $n$-th configuration $\xv_n$ to a list of intermediate states $s_n \coloneqq (\tau(\xv_n), T_{i_n}^{(n)}, \av_n)$.
  \vspace{-1mm}
\end{itemize}
Note that multiple intermediate states could exist for an HP configuration $\xv_n$.
Algorithm~\ref{main:methods:alg:automatic-job-alloc-wrapper} presents the pseudocode of our wrapper.
Although the algorithm itself is nothing special, we need to make sure that multiple workers will not edit the same file at the same time.
Since usecases of our wrapper are not really limited to intra-process but could be inter-process synchronization, we use \texttt{fcntl} to safely acquire file locks.
Furthermore, the test coverage of our code on MacOS and Ubuntu is $95^+\%$ to validate the behavioral correctness of our package.

\begin{algorithm}[tb]
  \caption{Automatic Waiting Time Scheduling Wrapper (see Figure~\ref{main:methods:subfig:api} as well)}
  \label{main:methods:alg:automatic-job-alloc-wrapper}
  \begin{algorithmic}[1]
    \Function{Worker}{$\xv^{(N)}, \av^{(N)}$}
    \State{Get intermediate states from the file if intermediate states of $\xv^{(N)}$ exist}
    \If{evaluation can be restarted from the $n$-th observation}
    \State{$\tau^\prime \leftarrow \tau(\xv_{n} | \av_n)$}
    \Comment{Note that $s_n \coloneqq (\tau_n, T^{(n)}_{i_n}, \av_n)$ s.t. $T_{i_n}^{(n)} \leq T_p^{(N)} + t^{(N)}$ cannot be used}
    \Else \Comment{Restart is not supported when there are multiple fidelities}
    \State{$\tau^\prime \leftarrow 0$}
    \label{main:methods:line:continual-eval}
    \EndIf
    \State{Query the result $r^{(N)} \coloneqq (f^{(N)}, \tau^{(N)}) \coloneqq (f(\xv^{(N)}|\av^{(N)}),\tau(\xv^{(N)}|\av^{(N)}))$}
    \State{$T_{\mathrm{now}} \leftarrow \max(T_{\mathrm{now}}, T_p^{(N)}) + t^{(N)}, T_p^{(N + 1)} \leftarrow T_{\mathrm{now}} + \tau^{(N)} - \tau^\prime, T_{p^\prime}^{(N + 1)} \leftarrow T_{p^\prime}^{(N)}~(p^\prime \neq p)$}
    \LineComment{$k \in \mathbb{Z}_{\geq 0}$ is the number of results from the other workers appended during the wait}
    \State{Wait till $p \in \argmin_{p \in [P]} T_p^{(N + k + 1)}$ satisfies}
    \State{Record (or update the state $s_n$ to) $s_{N+k+1} = (\tau^{(N)}, T_p^{(N +k+ 1)}, \av^{(N)})$ with a key $\xv^{(N)}$}
    \State{\textbf{return} $f_{N+k+1} \coloneqq f^{(N)}$}
    \EndFunction
    \Statex{$\D_0 \leftarrow \emptyset, T_p^{(0)} \leftarrow 0, T_{\mathrm{now}} \leftarrow 0, N \leftarrow 0$, and user needs to provide an optimizer policy $\pi$}
    \While{the budget is left}
    \Comment{This codeblock is run by $P$ different workers in parallel}
    \LineComment{e.g. BOHB and DEHB define a fidelity (in $\av^{(N)}$) at each iteration by themselves}
    \State{Sample $\xv^{(N)} \sim \pi(\cdot | \D_{N})$ with $t^{(N)}$ seconds and get $\av^{(N)}$ from a user-defined algorithm}
    \State{$f^{(N)} \leftarrow$ \texttt{worker}($\xv^{(N)}, \av^{(N)}$)}
    \LineComment{Recall that $\D_{\cdot}$ is modified from the other workers as well}
    \State{$\D_{N+k+1} \leftarrow \D_{N+k} \cup \{(\xv^{(N)}, f^{(N)})\}, N\leftarrow N+k+1$}
    \EndWhile
  \end{algorithmic}
\end{algorithm}

\section{Tool Usage}
In our package, \texttt{ObjectiveFuncWrapper} is the main module and it provides three different options:
(1) function wrapper for intra-process synchronization (e.g. DEHB and SMAC3),
(2) function wrapper for inter-process synchronization (e.g. NePS~\footnote{\url{https://github.com/automl/neps/}}), and
(3) function and optimizer wrapper for the ask-and-tell interface.
In this paper, we discuss the usage of Options (1) and (2), and we kindly ask users to refer to Appendix~\ref{appx:wrapper-ask-and-tell:section} for Option (3).

An instance of \texttt{ObjectiveFuncWrapper} serves as an objective function and we just need to pass it to an optimizer;
see Appendix~\ref{appx:wrapper-args:section} for the arguments of \texttt{ObjectiveFuncWrapper}.
When optimizers take a different interface, we can easily modify the interface via inheritance:

\begin{minipage}{0.95\linewidth}
\begin{lstlisting}[caption=An example of inheritance for a different interface.\label{main:usage:code:inheritance}]
class MyObjectiveFuncWrapper(ObjectiveFuncWrapper):
    def __call__(self, config, budget):
        # modify config into dict[str, Any]
        return super().__call__(
            eval_config=config,
            fidels={self.fidel_keys[0]: budget}
        )
\end{lstlisting}
\end{minipage}
In Listing~\ref{main:usage:code:inheritance}, the optimizer requires the objective function to have arguments named \texttt{config} instead of \texttt{eval\_config} and \texttt{budget} instead of \texttt{fidels}.
Then we need to somehow modify \texttt{config} into the format of \texttt{eval\_config}~(\texttt{dict[str, Any]}) if \texttt{config} is not \texttt{dict[str, Any]}.
We can also easily deactivate the MFO setting if \texttt{fidel\_keys=None} is specified:

\begin{minipage}{0.95\linewidth}
\begin{lstlisting}[caption=An example of inheritance for non MFO setting.\label{main:usage:code:non-mfo}]
class MyObjectiveFuncWrapper(ObjectiveFuncWrapper):
    def __call__(self, eval_config):
        # fidel_keys must be None; otherwise, get an error
        return super().__call__(eval_config=eval_config)
\end{lstlisting}
\end{minipage}
Further examples for BOHB, DEHB, NePS, and SMAC3 on MLP in Table~6 of HPOBench~(\cite{eggensperger2021hpobench}), HPOlib~(\cite{klein2019tabular}), JAHS-Bench-201~(\cite{bansal2022jahs}), LCBench~(\cite{zimmer2021auto}) in YAHPOBench~(\cite{pfisterer2022yahpo}), some synthetic functions.
See Appendices~\ref{appx:benchmarks:section},\ref{appx:optimizers:section} for more details.
The examples are available at
\ifunderreview
\url{https://anonymous.4open.science/r/mfhpo-simulator-3C81/}.
\else
\url{https://github.com/nabenabe0928/mfhpo-simulator/}.
\fi

\begin{figure}[t]
  \centering
  \includegraphics[width=0.98\textwidth]{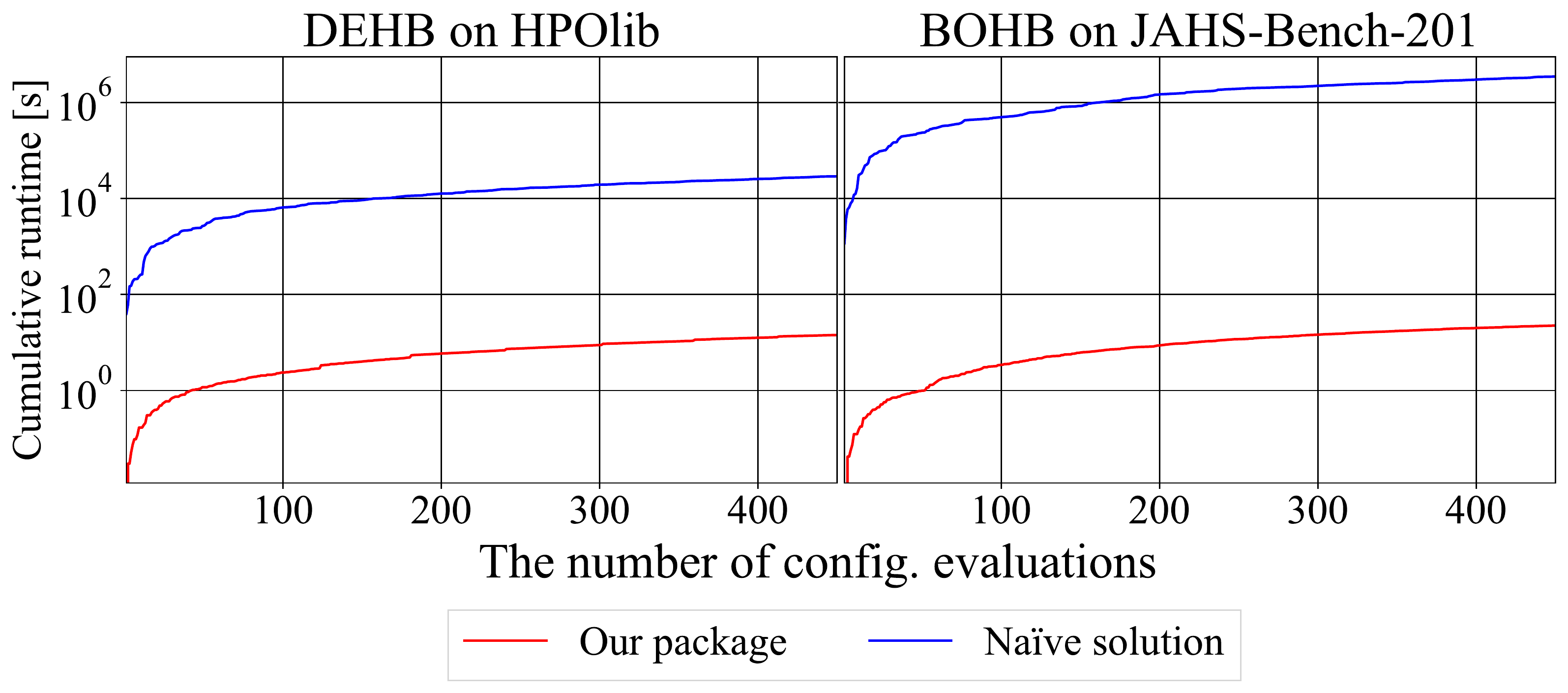}
  \vspace{-1mm}
  \caption{
    The benchmarks of runtime using our package (the red lines) and the na\"ive simulation (the blue lines), which waits for $\tau(\xv)$ at each iteration.
    Our benchmarks were performed on a machine with $12$ cores of Intel core i7-1255U.
    \textbf{Left}: the benchmark using DEHB~(\protect\cite{awad2021dehb}) on Slice Localization of HPOlib~(\protect\cite{klein2019tabular}).
    \textbf{Right}: the benchmark using BOHB~(\protect\cite{falkner2018bohb}) on CIFAR10 of JAHS-Bench-201~(\protect\cite{bansal2022jahs}).
    While BOHB took $23$ seconds and DEHB took $15$ seconds to finish each benchmark with our package, their na\"ive solutions took hours to days to finish benchmarks.
  }
  \label{main:experiments:fig:runtime-benchmark}
\end{figure}

\section{Broader Impact \& Limitations}
\label{main:broad-impact:section}

The primary motivation for this paper is to reduce the runtime of simulations for MFO.
In Figure~\ref{main:experiments:fig:runtime-benchmark}, we show an example of MFO using DEHB on Slice Localization of HPOlib~(\cite{klein2019tabular}) and BOHB on CIFAR10 of JAHS-Bench-201~(\cite{bansal2022jahs}).
Both examples use $\eta = 3$, which is a control parameter of HyperBand~(\cite{li2017hyperband}) and $P = 4$.
As can be seen, while we could finish $450$ evaluations in $15 \sim 25$ seconds with our wrapper, the na\"ive implementation required hours to days.
As a computer consumes about 30Wh even just for waiting and $736~\mathrm{g}$ of $\mathrm{CO}_2$ is produced per 1kWh, the experiment of BOHB on JAHS-Bench-201 would have produced about $6~\mathrm{kg}$ of $\mathrm{CO}_2$.
Therefore, researchers can reduce the corresponding amount of $\mathrm{CO}_2$ for each experiment.
The main limitation of our current wrapper is the assumption that none of the workers will not die and any additional workers will not be added after the initialization.
Besides that, our package cannot be used on Windows OS because \texttt{fcntl} is not supported on Windows.

\section{Conclusions}
In this paper, we presented a Python wrapper to maintain the exact order of the observations without waiting for actual runtimes.
Although some existing packages internally support the similar mechanism, they are not applicable to multiprocessing or multithread setups and they cannot be immediately used for newly developed methods.
On the other hand, our package supports such distributed computing setups and researchers can simply wrap their objective functions by our wrapper and they can directly give to their optimizers.
We describe the basic usage of our package and demonstrated that our package significantly reduce the $\mathrm{CO}_2$ production that experiments using tabular and surrogate benchmarks would have caused.

\bibliographystyle{apalike}
\bibliography{ref}

\ifappendix
\clearpage
\appendix
\section{Wrapper Object (\texttt{ObjectiveFuncWrapper})}
\label{appx:wrapper:section}
In this section, we describe more details on our wrapper.

\subsection{Arguments}
\label{appx:wrapper-args:section}
The arguments of \texttt{ObjectiveFuncWrapper} object are as follows:
\begin{itemize}
  \vspace{-1.5mm}
  \item \texttt{obj\_func}: the objective function that takes (\texttt{eval\_config}, \texttt{fidels}, \texttt{seed}, \texttt{**data\_to\_scatter}) as arguments and returns $f(\xv|\av)$ and $\tau(\xv|\av)$ where \texttt{eval\_config} is \texttt{dict[str, Any]},
  \vspace{-1.5mm}
  \item \texttt{launch\_multiple\_wrappers\_from\_user\_side (bool)}: whether to use intra-process~(\texttt{False}) or inter-process~(\texttt{True}) synchronization,
  \vspace{-1.5mm}
  \item \texttt{ask\_and\_tell (bool)}:
  whether to use simulator for ask-and-tell interface (\texttt{True}) or not,
  \vspace{-1.5mm}
  \item \texttt{save\_dir\_name (str | None)}: the results and the required information will be stored in \texttt{mfhpo-simulator-info/<save\_dir\_name>/},
  \vspace{-1.5mm}
  \item \texttt{n\_workers (int)}: the number of parallel workers $P$,
  \vspace{-1.5mm}
  \item \texttt{n\_evals (int)}: the number of evaluations to get,
  \vspace{-1.5mm}
  \item \texttt{n\_actual\_evals\_in\_opt (int)}: the number of HP configurations to be evaluated in an optimizer which is used only for checking if no hang happens (should take \texttt{n\_evals + n\_workers}),
  \vspace{-1.5mm}
  \item \texttt{continual\_max\_fidel (int | None)}:  when users would like to restart each evaluation from intermediate states, the maximum fidelity value for the target fidelity must be provided.
  Note that the restart is allowed only if there is only one fidelity parameter.
  If \texttt{None}, no restart happens as in Line~\ref{main:methods:line:continual-eval} of Algorithm~\ref{main:methods:alg:automatic-job-alloc-wrapper},
  \vspace{-1.5mm}
  \item \texttt{runtime\_key (str)}: the key of the runtime in the returned value of \texttt{obj\_func},
  \vspace{-1.5mm}
  \item \texttt{obj\_keys (list[str])}: the keys of the objective and constraint names in the returned value of \texttt{obj\_func} and the values of the specified keys will be stored in the result file,
  \vspace{-1.5mm}
  \item \texttt{fidel\_keys (list[str] | None)}: the keys of the fidelity parameters in input \texttt{fidels},
  \vspace{-1.5mm}
  \item \texttt{seed (int | None)}: the random seed to be used in the wrapper,
  \vspace{-1.5mm}
  \item \texttt{max\_waiting\_time (float)}: the maximum waiting time for each worker and if each worker did not get any update for this amount of time, it will return \texttt{inf},
  \vspace{-1.5mm}
  \item \texttt{store\_config (bool)}: whether to store configurations, fidelities, and seed used for each evaluation, and
  \vspace{-1.5mm}
  \item \texttt{check\_interval\_time (float)}: how often each worker should check whether a new job can be assigned to it.
  \vspace{-1.5mm}
\end{itemize}
Note that \texttt{data\_to\_scatter} is especially important when an optimizer uses multiprocessing packages such as \texttt{dask} or \texttt{multiprocessing}, which deserialize \texttt{obj\_func} every time we call.
By passing large-size data via \texttt{data\_to\_scatter}, the time for (de)serialization will be negligible if optimizers use \texttt{dask.scatter} or something similar internally.
We kindly ask readers to check any updates to the arguments at
\ifunderreview
\url{https://anonymous.4open.science/r/mfhpo-simulator-3C81/README.md}.
\else
\url{https://github.com/nabenabe0928/mfhpo-simulator/}.
\fi

\subsection{Wrapper for Ask-and-Tell Interface}
\label{appx:wrapper-ask-and-tell:section}
When optimizers take ask-and-tell interface, simulations can be run on a single worker while preserving the results, and hence simulations can be further accelerated.
Note that the bottleneck of simulations is the waiting time due to the communication among each worker and simulations on a single worker can address this problem.
The benefits of this option are (1) faster, (2) memory-efficient, and (3) stable.
On the other hand, the downsides are that (1) this option forces optimizers to have the ask-and-tell interface and (2) simulations may fail to precisely consider a bottleneck caused by parallel execution of expensive optimizers.
For more details, see \url{https://github.com/nabenabe0928/mfhpo-simulator/}.

\section{Benchmarks}
\label{appx:benchmarks:section}

We first note that since the Branin and the Hartmann functions must be minimized, our functions have different signs from the prior literature that aims to maximize objective functions and when $\zv = [z_1, z_2, \dots, z_K] \in \mathbb{R}^K$, our examples take $\zv = [z, z, \dots,z] \in \mathbb{R}^K$.
However, if users wish, users can specify $\zv$ as $\zv = [z_1, z_2, \dots, z_K]$ from \texttt{fidel\_dim}.

\subsection{Branin Function}

The Branin function is the following $2D$ function that has $3$ global minimizers and no local minimizer:
\begin{equation}
  \begin{aligned}
    f(x_1, x_2)
    = a (x_2 - b x_1^2 +cx_1 - r)^2 + s (1 - t) \cos x_1 + s
  \end{aligned}
\end{equation}
where $\xv \in [-5, 10]\times [0, 15]$, $a = 1$, $b = 5.1 / (4\pi^2)$, $c = 5/\pi$, $r = 6$, $s=10$, and $t = 1/(8\pi)$.
The multi-fidelity Branin function was invented by \cite{kandasamy2020tuning} and it replaces $b, c, t$ with the following $b_{\zv}, c_{\zv}, t_{\zv}$:
\begin{equation}
  \begin{aligned}
    b_{\zv} & \coloneqq b - \delta_b (1 - z_1),               \\
    c_{\zv} & \coloneqq c - \delta_c (1 - z_2), \mathrm{~and} \\
    t_{\zv} & \coloneqq t + \delta_t (1 - z_3),               \\
  \end{aligned}
\end{equation}
where $\zv \in [0, 1]^3$, $\delta_b = 10^{-2}$, $\delta_c = 10^{-1}$, and $\delta_t = 5 \times 10^{-3}$.
$\delta_{\cdot}$ controls the rank correlation between low- and high-fidelities and higher $\delta_{\cdot}$ yields less correlation.
The runtime function for the multi-fidelity Branin function is computed as~\footnote{
  See the implementation of \cite{kandasamy2020tuning}: \texttt{branin\_mf.py} at \url{https://github.com/dragonfly/dragonfly/}.
}:
\begin{equation}
  \begin{aligned}
    \tau(\zv) \coloneqq C (0.05 + 0.95 z_1^{3/2})
  \end{aligned}
\end{equation}
where $C \in \mathbb{R}_+$ defines the maximum runtime to evaluate $f$.

\subsection{Hartmann Function}
The following Hartmann function has $4$ local minimizers for the $3D$ case and $6$ local minimizers for the $6D$ case:
\begin{equation}
  \begin{aligned}
    f(\xv) \coloneqq -\sum_{i=1}^{4} \alpha_i \exp \biggl[
      - \sum_{j=1}^3 A_{i,j} (x_j - P_{i,j})^2
      \biggr]
  \end{aligned}
\end{equation}
where $\boldsymbol{\alpha} = [1.0, 1.2, 3.0, 3.2]^\top$, $\xv \in [0, 1]^D$, $A$ for the $3D$ case is
\begin{equation}
  A = \begin{bmatrix}
    3   & 10 & 30 \\
    0.1 & 10 & 35 \\
    3   & 10 & 30 \\
    0.1 & 10 & 35 \\
  \end{bmatrix},
\end{equation}
$A$ for the $6D$ case is
\begin{equation}
  A = \begin{bmatrix}
    10   & 3   & 17   & 3.5 & 1.7 & 8  \\
    0.05 & 10  & 17   & 0.1 & 8   & 14 \\
    3    & 3.5 & 1.7  & 10  & 17  & 8  \\
    17   & 8   & 0.05 & 10  & 0.1 & 14 \\
  \end{bmatrix},
\end{equation}
$P$ for the $3D$ case is
\begin{equation}
  P = 10^{-4} \times \begin{bmatrix}
    3689 & 1170 & 2673 \\
    4699 & 4387 & 7470 \\
    1091 & 8732 & 5547 \\
    381  & 5743 & 8828 \\
  \end{bmatrix},
\end{equation}
and $P$ for the $6D$ case is
\begin{equation}
  P = 10^{-4} \times \begin{bmatrix}
    1312 & 1696 & 5569 & 124  & 8283 & 5886 \\
    2329 & 4135 & 8307 & 3736 & 1004 & 9991 \\
    2348 & 1451 & 3522 & 2883 & 3047 & 6650 \\
    4047 & 8828 & 8732 & 5743 & 1091 & 381  \\
  \end{bmatrix}.
\end{equation}
The multi-fidelity Hartmann function was invented by \cite{kandasamy2020tuning} and it replaces $\boldsymbol{\alpha}$ with the following $\boldsymbol{\alpha}_{\zv}$:

\begin{equation}
\begin{aligned}
  \boldsymbol{\alpha}_{\zv} \coloneqq \delta (1 - \zv)
\end{aligned}
\end{equation}
where $\zv \in [0,1]^4$ and $\delta = 0.1$ is the factor that controls the rank correlation between low- and high-fidelities.
Higher $\delta$ yields less correlation.
The runtime function of the multi-fidelity Hartmann function is computed as~\footnote{
  See the implementation of \cite{kandasamy2020tuning}: \texttt{hartmann3\_2\_mf.py} for the $3D$ case and \texttt{hartmann6\_4\_mf.py} for the $6D$ case at \url{https://github.com/dragonfly/dragonfly/}.
}:
\begin{equation}
\begin{aligned}
  \tau(\zv) &= \frac{1}{10} + \frac{9}{10}\frac{z_1 + z_2^3 + z_3 z_4}{3} \\
\end{aligned}
\end{equation}
for the $3D$ case and 
\begin{equation}
\begin{aligned}
  \tau(\zv) &= \frac{1}{10} + \frac{9}{10}\frac{z_1 + z_2^2 + z_3 + z_4^3}{4} \\
\end{aligned}
\end{equation}
for the $6D$ case where $C \in \mathbb{R}_+$ defines the maximum runtime to evaluate $f$.

\subsection{Tabular \& Surrogate Benchmarks}
In this paper, we used the MLP benchmark in Table~6 of HPOBench~(\cite{eggensperger2021hpobench}), HPOlib~(\cite{klein2019tabular}), JAHS-Bench-201~(\cite{bansal2022jahs}), and LCBench~(\cite{zimmer2021auto}) in YAHPOBench~(\cite{pfisterer2022yahpo}).

HPOBench is a collection of tabular, surrogate, and raw benchmarks.
In our example, we have the MLP (multi-layer perceptron) benchmark, which is a tabular benchmark, in Table~6 of the HPOBench paper~(\cite{eggensperger2021hpobench}).
This benchmark has $8$ classification tasks and provides the validation accuracy, runtime, F1 score, and precision for each configuration at epochs of $\{3, 9, 27, 81, 243\}$.
The search space of MLP benchmark in HPOBench is provided in Table~\ref{appx:benchmarks:tab:hpobench}.

HPOlib is a tabular benchmark for neural networks on regression tasks (Slice Localization, Naval Propulsion, Protein Structure, and Parkinsons Telemonitoring).
This benchmark has $4$ regression tasks and provides the number of parameters, runtime, and training and validation mean squared error (MSE) for each configuration at each epoch.
The search space of HPOlib is provided in Table~\ref{appx:benchmarks:tab:hpolib}.

JAHS-Bench-201 is an XGBoost surrogate benchmark for neural networks on image classification tasks (CIFAR10, Fashion-MNIST, and Colorectal Histology).
This benchmark has $3$ image classification tasks and provides FLOPS, latency, runtime, architecture size in megabytes, test accuracy, training accuracy, and validation accuracy for each configuration with two fidelity parameters: image resolution and epoch.
The search space of JAHS-Bench-201 is provided in Table~\ref{appx:benchmarks:tab:jahs}.

LCBench is a random-forest surrogate benchmark for neural networks on OpenML datasets.
This benchmark has $34$ tasks and provides training/test/validation accuracy, losses, balanced accuracy, and runtime at each epoch.
The search space of HPOlib is provided in Table~\ref{appx:benchmarks:tab:lc}.

\begin{table}[t]
  \begin{center}
    \caption{
      The search space of the MLP benchmark in HPOBench ($5$ discrete + $1$ fidelity parameters).
      Note that we have $2$ fidelity parameters only for the raw benchmark.
      Each benchmark has performance metrics of $30000$ possible configurations with $5$ random seeds.
    }
    \label{appx:benchmarks:tab:hpobench}
    \begin{tabular}{ll}
      \toprule
      Hyperparameter              & Choices                                                                              \\
      \midrule
      L2 regularization                  & [$10^{-8}, 1.0$] with $10$ evenly distributed grids                                                             \\
      Batch size                  & [$4, 256$] with $10$ evenly distributed grids                                                             \\
      Initial learning rate       & [$10^{-5}, 1.0$] with $10$ evenly distributed grids \\
      Width     & [$16, 1024$] with $10$ evenly distributed grids                                                     \\
      Depth        & \{$1,2,3$\}                                                                    \\
      \midrule
      Epoch~(\textbf{Fidelity})            & \{$3,9,27,81,243$\}                                                                           \\
      \bottomrule
    \end{tabular}
  \end{center}
\end{table}

\begin{table}[t]
  \begin{center}
    \caption{
      The search space of HPOlib ($6$ discrete + $3$ categorical + $1$ fidelity parameters).
      Each benchmark has performance metrics of $62208$
      possible configurations with $4$ random seeds.
    }
    \label{appx:benchmarks:tab:hpolib}
    \begin{tabular}{ll}
      \toprule
      Hyperparameter              & Choices                                                                              \\
      \midrule
      Batch size                  & \{$2^3, 2^4, 2^5, 2^6$\}                                                             \\
      Initial learning rate       & \{$5 \times 10^{-4}, 10^{-3}, 5 \times 10^{-3}, 10^{-2},5 \times 10^{-2}, 10^{-1}$\} \\
      Number of units \{1,2\}     & \{$2^4, 2^5, 2^6, 2^7, 2^8, 2^9$\}                                                   \\
      Dropout rate \{1,2\}        & \{$0.0,0.3,0.6$\}                                                                    \\
      \midrule
      Learning rate scheduler     & \{\texttt{cosine}$,$ \texttt{constant}\}                                             \\
      Activation function \{1,2\} & \{\texttt{relu}$,$ \texttt{tanh}\}                                                   \\
      \midrule
      Epoch~(\textbf{Fidelity})            & [$1, 100$]                                                                           \\
      \bottomrule
    \end{tabular}
  \end{center}
\end{table}

\begin{table}[t]
  \begin{center}
    \caption{
      The search space of JAHS-Bench-201 ($2$ continuous + $2$ discrete + $8$ categorical + $2$ fidelity parameters).
      JAHS-Bench-201 is an XGBoost surrogate benchmark and the outputs are deterministic.
    }
    \label{appx:benchmarks:tab:jahs}
    \begin{tabular}{lll}
      \toprule
      Hyperparameter                                      & Range or choices                                              \\
      \midrule
      Learning rate                                       & $[10^{-3}, 1]$                                                \\
      L2 regularization                                   & $[10^{-5}, 10^{-2}]$                                          \\
      Activation function                                 & \{\texttt{ReLU}, \texttt{Hardswish}, \texttt{Mish}\}          \\
      Trivial augment~(\cite{muller2021trivialaugment})   & \{\texttt{True}, \texttt{False}\}                             \\
      \midrule
      Depth multiplier                                    & $\{1, 2, 3\}$                                                 \\
      Width multiplier                                    & $\{2^2, 2^3, 2^4\}$                                           \\
      \midrule
      Cell search space                                   & \{\texttt{none}, \texttt{avg-pool-3x3}, \texttt{bn-conv-1x1}, \\
      (NAS-Bench-201~(\cite{dong2020bench}), Edge 1 -- 6) & \: \texttt{bn-conv-3x3}, \texttt{skip-connection}\}         \\
      \midrule
      Epoch~(\textbf{Fidelity})                                    & [$1, 200$]                                                    \\
      Resolution~(\textbf{Fidelity})                               & [$0.0, 1.0$]                                                    \\
      \bottomrule
    \end{tabular}
  \end{center}
\end{table}

\begin{table}[t]
  \begin{center}
    \caption{
      The search space of HPOlib ($3$ discrete + $4$ continuous + $1$ fidelity parameters).
      Although the original LCBench is a collection of $2000$ random configurations, YAHPOBench created random-forest surrogates over the $2000$ observations.
      Users can choose deterministic or non-deterministic outputs.
    }
    \label{appx:benchmarks:tab:lc}
    \begin{tabular}{ll}
      \toprule
      Hyperparameter        & Choices              \\
      \midrule
      Batch size            & [$2^4, 2^9$]         \\
      Max number of units   & [$2^6, 2^{10}$]      \\
      Number of layers      & [$1, 5$]             \\
      \midrule
      Initial learning rate & [$10^{-4}, 10^{-1}$] \\
      L2 regularization     & [$10^{-5}, 10^{-1}$] \\
      Max dropout rate      & [$0.1, 0.9$]         \\
      Momentum              & [$0.1, 0.9$]         \\
      \midrule
      Epoch~(\textbf{Fidelity})                 & [$1, 52$]           \\
      \bottomrule
    \end{tabular}
  \end{center}
\end{table}

\section{Optimizers}
\label{appx:optimizers:section}
In our package, we show examples using BOHB~(\cite{falkner2018bohb}), DEHB~(\cite{awad2021dehb}), SMAC3~(\cite{lindauer2022smac3}), and NePS~\footnote{\url{https://github.com/automl/neps/}}.
BOHB is a combination of HyperBand~(\cite{li2017hyperband}) and tree-structured Parzen estimator~(\cite{bergstra2011algorithms,watanabe2023tree}).
DEHB is a combination of HyperBand and differential evolution.
SMAC3 is an HPO framework.
SMAC3 supports various Bayesian optimization algorithms and uses different strategies for different scenarios.
The default strategies for MFO is the random forest-based Bayesian optimization and HyperBand.
NePS is another HPO framework jointly with neural architecture search.

\else

\customlabel{data1}{1}
\customlabel{data2}{2}

\fi

\end{document}